\documentclass[runningheads]{llncs}
\usepackage[T1]{fontenc}
\usepackage{amsmath}
\usepackage{amssymb}
\usepackage{xcolor}
\usepackage{hyperref}
\usepackage{multirow}
\usepackage{subcaption}
\usepackage{booktabs} 
\usepackage{float}
\usepackage{graphicx}
\usepackage{svg}

\begin{document}
\title{TAG‑Head: Time‑Aligned Graph Head for Plug-and-Play Fine‑grained Action Recognition}
\titlerunning{TAG‑Head}
%
\author{Imtiaz Ul Hassan\orcidID{0009-0003-3568-3825}\and Nik Bessis\orcidID{0000-0002-6013-3935} \and
Ardhendu Behera \orcidID{0000-0003-0276-9000} 
}
\authorrunning{Hassan et al.}

\institute{
Department of Computer Science, Edge Hill University \\ Ormskirk, L39 4QP, United Kingdom \\
\email{\{hassani, bessisn, beheraa\}@edgehill.ac.uk} \\
}

\maketitle              
\begin{abstract}
Fine-grained human action recognition (FHAR) is challenging because visually similar actions differ by subtle spatio-temporal cues. Many recent systems enhance discriminability with extra modalities (e.g., pose, text, optical flow), but this increases annotation burden and computational cost. We introduce \textbf{TAG-Head}, a lightweight spatio-temporal graph head that upgrades standard 3D backbones (SlowFast, R(2+1)D-34, I3D, etc.) for FHAR using \emph{RGB only}. Our pipeline first applies a \emph{Transformer encoder} with learnable 3D positional encodings to the backbone tokens, capturing long-range dependencies across space and time. The resulting features are then refined by a graph in which (i) \emph{fully-connected intra-frame} edges to resolve subtle appearance differences within frames, and (ii) \emph{time-aligned temporal} edges that connect features at the \emph{same spatial location} across frames to stabilise motion cues without over‑smoothing. The head is compact (little parameter/FLOP overhead), plug-and-play across backbones, and trained end-to-end with the backbone. Extensive evaluations on FineGym (Gym99 and Gym288) and HAA500 show that TAG-Head sets a new state-of-the-art among RGB-only models and surpasses many recent multimodal approaches (video + pose + text) that rely on privileged information. Ablations disentangle the contributions of the Transformer and the graph topology, and complexity analyses confirm low latency. TAG-Head advances FHAR by explicitly coupling global context with high-resolution spatial interactions and low-variance temporal continuity inside a slim, composable graph head. The simplicity of the design enables straightforward adoption in practical systems that favour RGB‑only sensors, while delivering performance gains typically associated with heavier or multimodal models. \emph{Code will be released on GitHub.}


\end{abstract}
\section{Introduction }
\begin{figure}[t]
    \centering
    \begin{minipage}{0.38\linewidth}
        \centering
        \includegraphics[width=\linewidth]{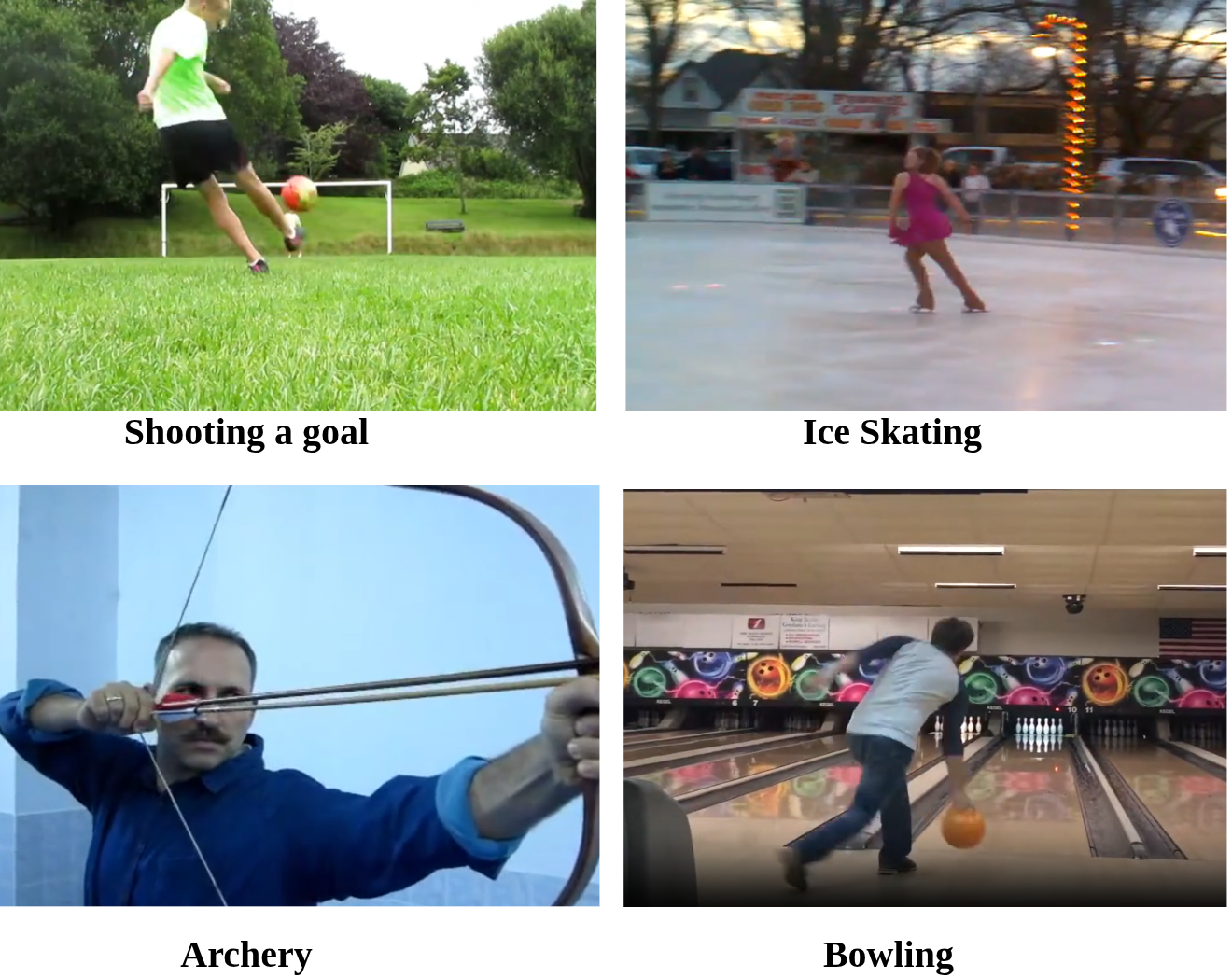}
        \caption*{(a) Coarse-grained actions}
    \end{minipage}
    \hfill
    \begin{minipage}{0.60\linewidth}
        \centering
        \includegraphics[width=\linewidth]{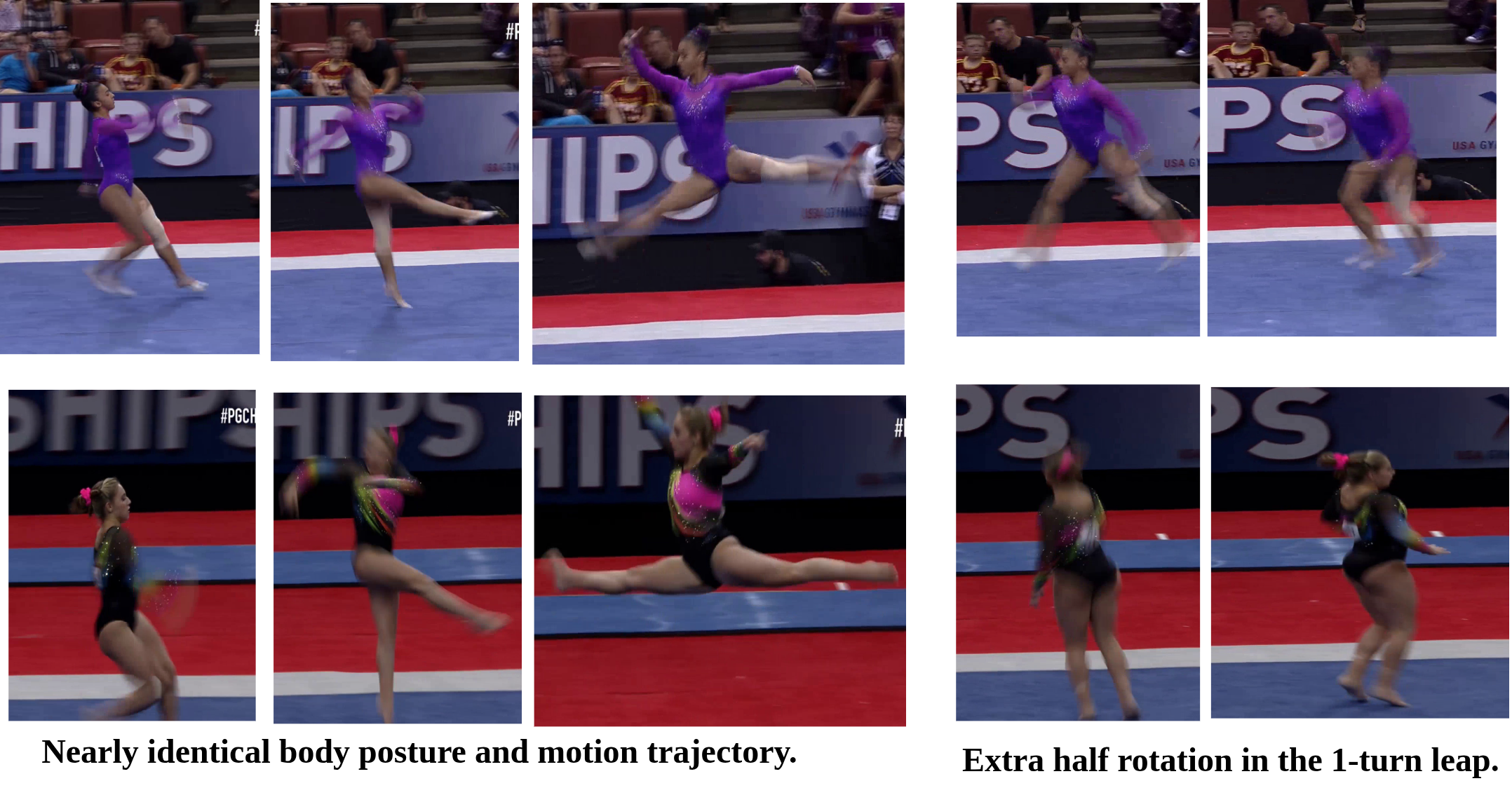}
        \caption*{(b) Fine-grained actions}
    \end{minipage}
    
    \vspace{2mm} 
    
    \caption{
        \textbf{Visual comparison between coarse-grained and fine-grained action classes.} 
        (a) Coarse-grained actions such as Shooting a Goal, Ice Skating, Archery, and Bowling exhibit distinct global motion patterns and environments. 
        (b) Fine-grained actions, specifically ``Switch leap with 0.5 turn'' vs. ``Switch leap with 1 turn'', demonstrate nearly identical visual contexts where only subtle differences in rotation distinguish the classes, highlighting the challenge of fine-grained classification.
    }
    \label{fig1}
\end{figure}

Fine-grained human action recognition (FHAR) plays a crucial role in sports analytics, surveillance, and HCI, where outcomes rely on subtle, localised variations instead of broad visual appearance or large-scale movements. Unlike coarse actions (e.g., Ice Skating, Archery; Fig.~\ref{fig1}(a)), fine-grained classes differ by nuanced spatio-temporal dynamics—micro-motions, phase shifts, or small angular differences. In FineGym99, for example, ``switch leap with 0.5 turn'' vs. ``switch leap with 1 turn'' are visually similar and primarily separable by rotational extent, demanding precise temporal modelling (Figs.~\ref{fig1}(b)). These challenges motivate architectures that preserve high-resolution spatial detail while integrating context over longer time horizons.

Early progress in FHAR largely strengthened RGB pipelines maturing 3D CNNs such as I3D \cite{i3d2017}, R(2+1)D \cite{r2plus1d}, and SlowFast \cite{slowfast2019}, as well as ViT-style video transformers \cite{fan2021multiscale,arnab2021vivit},  while the community consolidated evaluation on fine-grained datasets such as FineGym \cite{finegym} (hierarchical gymnastics) and HAA500 \cite{haa500}  (atomic human actions). These benchmarks emphasise subtle intra-class variation and catalyse research on higher temporal precision. A parallel trend leverages \emph{additional modalities}. Pose-guided models couple skeleton streams with video transformers to inject strong kinematic priors, reporting consistent gains on fine-grained sport settings, e.g., PGVT \cite{pgvt2024}. Vision-language (VL) models extended to video carry over textual semantics and layered descriptions as supervisory signals; more recent approaches further enrich VL backbones with pose information to better capture fine-grained differences \cite{pangea2024,pevl2024}. While these methods work well, they come with a deployment overhead: (i) reliance on high-quality pose estimators at test time (computational overhead; failure under occlusion/camera motion) and (ii) cross-modal fusion costs and sensitivity to prompt/annotation design in VL pipelines. Beyond RGB+pose and VL, other modalities such as 3D point clouds ~\cite{ben20243dinaction} and video-LLMs \cite{chen2024videollm} have been explored for fine-grained temporal understanding. However, they typically demand heavier compute or specialised sensors, limiting adoption in standard RGB-only settings.

In many real deployments (broadcast sports, CCTV, consumer devices), only RGB is available or desirable for cost, latency, and privacy reasons. Recent works \cite{yang2024adapting} show that improving \emph{temporal range} and \emph{tokenization} in video transformers can narrow the gap to multimodal models, but maintaining fine-grained spatial fidelity remains challenging as token budgets are constrained. This motivates architectures that can (i) capture long-range dependencies to resolve near-duplicate classes and (ii) explicitly reason about high-resolution spatial interactions without incurring quadratic attention costs everywhere.

\noindent\textbf{Motivation and contribution:}
We argue that FHAR benefits from a \emph{hybrid} design that cleanly separates global context modelling from local topology-aware refinement. Concretely, we introduce \textbf{TAG-Head}, a plug-and-play head for standard 3D backbones that first applies a compact Transformer encoder with \emph{learnable 3D positional encodings} to encode long-range space–time relations, then performs graph-based refinement using a \emph{time-aligned} topology; fully connected spatial interactions \emph{within} each frame to sharpen fine contrasts, and strictly aligned connections \emph{across} frames at the same spatial site to stabilise motion cues. Instead of stacking deep parametric Graph Neural Network (GNN) layers (which risk oversmoothing on dense intra-frame cliques and add parameters), we adopt personalized PageRank propagation, which provides deep, controllable propagation while continuously anchoring features to the Transformer evidence via a teleport (restart) diffusion term. This RGB-only formulation recovers much of the discriminative power often attributed to extra modalities without the associated annotation, latency, and robustness costs, and achieves state-of-the-art results on fine-grained benchmarks such as FineGym and HAA500.
Our main contributions are:
\begin{enumerate}
    \item We position FHAR within the latest landscape of RGB-only and multimodal methods, highlighting the benefits and practical trade-offs of pose/VL extensions.

    \item We introduce a lightweight Transformer+Graph head that marries global context with explicit, high-resolution spatial interactions and low-variance temporal continuity.

    \item We conducted extensive experiments on three challenging FHAR benchmarks (\textit{FineGym99}, \textit{FineGym288}, and \textit{HAA500}), demonstrating a new state-of-the-art among RGB-only models and outperforming several recent multimodal methods. Detailed ablations and complexity analysis showing favourable accuracy/latency trade-offs relative to recent multimodal systems, making it practical for real-world deployment.
\end{enumerate}


\section{Related Work}

FHAR has evolved through multiple paradigms, beginning with RGB-only models and expanding into skeleton-based GNNs and multimodal fusion methods. Each paradigm contributes unique strengths and limitations. This section categorises prior work into three major areas: RGB-based approaches, GNNs for action modelling, and recent advancements in vision-language and multimodal frameworks.

\noindent \textbf{RGB-only Approaches:} RGB-based models serve as the cornerstone of FHAR, evolving from 2D CNN-based frameworks such as TSN~\cite{tsn2018} and TPN~\cite{tpn2020} to advanced 3D CNNs such as I3D~\cite{i3d2017}, R(2+1)D~\cite{r2plus1d}, SlowFast~\cite{slowfast2019}, and X3D~\cite{x3d}. These architectures, in addition to recent video transformers such as ViViT~\cite{arnab2021vivit}, TQN~\cite{tqn2021}, and MViT~\cite{fan2021multiscale}, have demonstrated strong general-purpose performance and offer the significant practical advantage of operating directly on raw pixels, bypassing the need for additional sensors. However, on recent fine-grained benchmarks such as FineGym~\cite{finegym} and HAA500~\cite{haa500}, which emphasise subtle motion and short temporal sequences, their performance is often limited. Key shortcomings include an over-reliance on scene context rather than fine-grained human motion, insufficient modelling of the precise long-range temporal dependencies needed to discriminate phases or micro-movements, and a limited capacity to resolve subtle spatiotemporal differences between visually similar actions~\cite{tqn2021}. This exposes a critical gap between general video understanding and the specific demands of FHAR. Our approach addresses this by introducing graph-based spatial refinement; by modelling intra-frame interactions as a graph, TAG-Head explicitly forces the network to reason about local topological relationships and nuanced micro-motions, effectively decoupling discriminative motion cues from static background noise.

\noindent \textbf{GNNs for Action Recognition:} GNNs~\cite{kipf2017semi} have become the standard for modelling relational structures in FHAR, particularly through skeleton-based Graph Convolutional Networks (GCNs)~\cite{gcn2018,as_gcn2019,2sagcn2019,shi2019skeleton} and their 2D-CNN adaptations~\cite{fgkgvl2024}. Recent research has pivoted toward hybrid architectures that merge graph convolutions with attention mechanisms. For example, SAE-GNN~\cite{geng2023saegnn} uses angular encoding and self-attention for fine-grained recognition, while STRIKE-net~\cite{strike2025} employs a spatiotemporal graph transformer to learn hierarchical representations for complex actions such as soccer. Furthermore, multi-aggregation and feature-gated attention strategies~\cite{humnabadkar2024driving} have shown promise in enhancing dynamic spatiotemporal modelling. 
Although these methods demonstrate strong relational reasoning capabilities, their reliance on explicit skeleton inputs obtained via pose estimation introduces potential errors from inaccurate pose extraction, increased computational costs, and information loss from discarding appearance features. Our approach addresses this limitation by incorporating GNN modules to capture spatial feature relations directly from RGB representations, maintaining the relational modelling advantages of graph architectures while eliminating the dependency on pose estimation.

\noindent \textbf{Vision-Language Models and Multimodal Approaches:} Beyond skeletal graphs, another strategy involves combining multiple modalities, particularly through vision-language (VL) models. Early multimodal methods~\cite{vilp2023,skeleton2022} incorporated human pose information alongside video features to capture structural joint relationships. With the advent of large-scale VL models such as CLIP~\cite{clip2021} and ALIGN~\cite{jia2021}, there has been a growing interest in integrating textual semantics into action recognition. ActionCLIP~\cite{actionclip2021} and XCLIP~\cite{xclip2022} pioneered aligning video content with textual descriptions to establish robust multimodal representation spaces. Recent works like Pangea~\cite{pangea2024}, PGVT~\cite{pgvt2024}, and PeVL~\cite{pevl2024} further incorporate pose into VL frameworks to improve recognition. Despite their power, multimodal approaches face practical challenges, including high computational costs and a heavy reliance on high-quality external annotations. These limitations underscore the value of improving RGB-only methods, which are more scalable and easier to deploy. Motivated by these challenges, we introduce TAG-Head, a lightweight module that integrates the relational reasoning of GNNs with the global dependency modelling of Transformers. This hybrid approach enables high-fidelity FHAR using only RGB data, effectively bypassing the computational overhead and annotation requirements of multimodal and pose-dependent frameworks.

\begin{figure}[t]
\centering
\includegraphics[width=\textwidth]{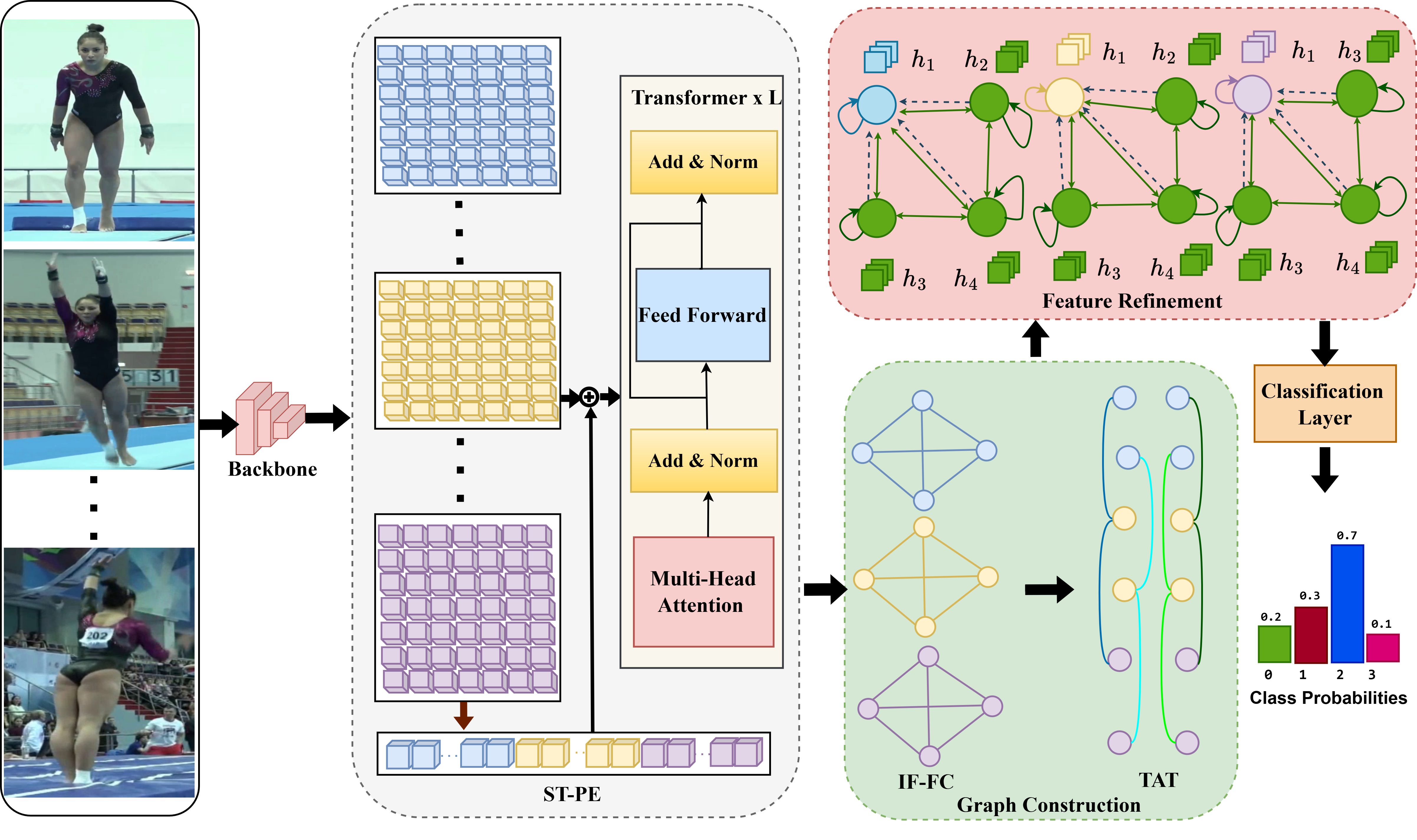}
\caption{
Overview of the proposed TAG-Head framework. The framework extracts spatio-temporal tokens from an input video using a 3D backbone. These tokens are enriched with learnable 3D positional encodings  (ST-PE) and processed by a Transformer encoder to model global dependencies. Subsequently, a spatio-temporal graph is constructed using intra-frame fully connected (IF-FC) and time-aligned temporal (TAT)  edges. The Feature Refinement Module utilises APPNP-based propagation to sharpen spatial contrasts and stabilise temporal cues, followed by global spatio-temporal average pooling for FHAR classification.}
\label{fig3}
\end{figure}

\section{Methodology}
An overview of our TAG-Head framework is shown in Fig.~\ref{fig3}. The architecture consists of three core components: (1) a standard 3D backbone to generate spatio-temporal tokens for a given input video; (2) a lightweight Transformer encoder with learnable 3D positional encodings for the spatio-temporal tokens to model long-range dependencies across space and time; and (3) our novel lightweight spatio-temporal graph head that refines features via dense \emph{intra-frame} interactions and \emph{time-aligned} propagation across frames. This hybrid design pairs the content-adaptive global context (Transformer) with topology-faithful diffusion (graph), which is precisely what FHAR requires; subtle spatial contrasts and temporally stable cues without heavy multimodal machinery.

\noindent \textbf{Problem formulation: }
Given a video clip, a 3D backbone $\mathcal{B}$ produces spatiotemporal features $\mathbf{F} \in \mathbb{R}^{T \times H \times W \times C}$. Let $P=W\times H$ and index spatial positions by $p\in \{1,\dots,P\}$ and frames by $t\in \{1,\dots ,T\}$. We represent the $C$-dimensional feature at $(t,p)$ by $\mathbf{x}_{t,p}\in \mathbb{R}^C$. We flatten the grid to a token sequence of length $N=T.P$ only within the Transformer block, then reshape back to the grid for graph refinement.

\noindent \textbf{Learnable 3D positional encoding: }
To give tokens with space–time identity, each $\mathbf{x}_{t,p}$ receives a learnable embedding $\mathbf{e}_{t,p}\in  \mathbb{R}^C$, producing $\mathbf{z}_{t,p}=\mathbf{x}_{t,p}+\mathbf{e}_{t,p}$.

\noindent \textbf{Transformer encoder (global long-range modelling): }
Let $\mathbf{Z}^{(l)} \in \mathbb{R}^{N \times C}$ be the input token matrix to the layer $\{l=0,\dots ,K-1\}$ with $\mathbf{Z}^{(0)} = \{\mathbf{z}_{t,p}\}$. For head $h=1,\dots ,H$ with per-head dimension $d_h=C/H$, query $\mathbf{Q}_h^{(l)}$, key $\mathbf{K}_h^{(l)}$ and value $\mathbf{V}_h^{(l)}$ for each head $h$ in layer $l$ are computed using linear projections:
\[
\mathbf{Q}^{(l)}_h = \mathbf{Z}^{(l)} \mathbf{W}_Q^{(h)}, \quad \mathbf{K}^{(l)}_h = \mathbf{Z}^{(l)} \mathbf{W}_K^{(h)}, \quad \mathbf{V}^{(l)}_h = \mathbf{Z}^{(l)} \mathbf{W}_V^{(h)},
\]
where projection matrices $\mathbf{W}_{\{Q,K,V\}}^{(h)}\in \mathbb{R}^{C\times d_h}$ and  are learnt for each head $h$. This is followed by the computation of the attention $\mathbf{A}_h^{(l)}$ for each head $h$ and the output projection at each layer $l$:
\begin{equation}
\mathbf{A}_h^{(l)} = \text{softmax}\left(\frac{\mathbf{Q}_h^{(l)} (\mathbf{K}_h^{(l)})^\top}{\sqrt{d_h}}\right)\mathbf{V}_h^{(l)}, \quad \text{MHA}(\mathbf{Z}^{(l)}) = \mathop{\Vert}_{h=1}^{H} \mathbf{A}_h^{(l)}\mathbf{W}_o
\end{equation}
\noindent with $\mathbf{W}_o\in \mathbb{R}^{(H d_h)\times C}$ and $\Vert$ denoting concatenation.  
We use pre-norm residual blocks to compute the final feature i.e. $\mathbf{U}^{(l)}=\mathbf{Z}^{(l)}+\text{MHA}(\text{LN}(\mathbf{Z}^{(l)}))$ and $\mathbf{Z}^{(l+1)}=\mathbf{U}^{(l)}+\text{MLP}(\text{LN}(\mathbf{U}^{(l)}))$, MLP is a multi-layer perceptron, and LN is the layer norm. We reshape $\mathbf{Z}^{(K)} = \mathbf{Z}^{(l+1)}$ to $\tilde{\mathbf{F}}\in \mathbb{R}^{T \times H \times W \times C}$ and denote refine tokens by $\tilde{\mathbf{z}}_{t,p}$ to the following graph stage as initial features.

\noindent\textbf{Graph construction and feature refinement: }
We treat each transformer-refined token $\tilde{\mathbf{z}}_{t,p}\in \mathbb{R}^C$ as a node and build a spatio-temporal graph $\mathcal{G} = (\mathcal{V}, \mathcal{E})$. The nodes are indexed by time $t$ and spatial location $p$; thus $|\mathcal{V}|=N=T.P$. The typed edge set $\mathcal{E}=\mathcal{E}_{\text{intra}}\cup \mathcal{E}_{\text{temp}}$. $\mathcal{E}_{\text{intra}}$ represents the fully-connected nodes in each frame $t$, i.e. $((t,p),(t,q))\in \mathcal{E}_{\text{intra}}$ $\forall p\ne q$. This induces a complete graph over the $P$ spatial locations for each frame $t$. Similarly, $\mathcal{E}_{\text{temp}}$ connects the nodes temporally, i.e. $((t,p),(t\pm 1,p))\in \mathcal{E}_{\text{temp}}$.
The aim is to propagate evidence (i) densely \textit{within} a frame (to sharpen fine contrasts) and (ii) only along the same spatial location over time (to stabilise motion), without heavy extra parameters or training instabilities. 

Let $M_{\text{intra}}$, $M_{\text{temp}}\in\{0,1\}^{N\times N}$ be the corresponding adjacency matrices (without self-loops). We form the combined adjacency matrix 
$M=\big(\begin{smallmatrix}
  M_{\text{intra}} & 0\\
  0 & M_{\text{temp}}
\end{smallmatrix}\big)$. We further introduce self-loops $\tilde{M}=M+I$ by adding the identity matrix $I$. This is required to update the node features using the GNN message passing mechanism between nodes. A standard layer-wise graph convolutional scheme \cite{kipf2017semi} expands the receptive field by stacking layers, but it faces two limitations: (i) the averaging inherent to message aggregation causes \textit{over-smoothing} when many layers are used, eroding local discriminability; and (ii) enlarging the effective neighborhood by adding layers inflates depth and parameters, since each layer introduces new learnable weights. To mitigate these issues, we adapt the approximate personalised propagation of neural predictions (APPNP) algorithm \cite{appnp2018}. It achieves linear computational complexity by approximating topic-sensitive PageRank through power iteration, which relates to a random walk with restarts. The $K_\text{prop}$ steps of personalised PageRank propagation with the teleport (restart) parameter $\alpha \in (0,1)$ is calculated as: 

\begin{equation}
\mathbf{H}^{(k+1)} = (1-\alpha) \mathcal{A} \mathbf{H}^{(k)} + \alpha \mathbf{H}^{(0)}, \quad \mathcal{A} = D^{-1/2}\tilde{M}D^{-1/2}, \quad \mathbf{H}^{(0)} = \tilde{\mathbf{F}}
\end{equation}

\noindent where $D$ is the diagonal node degree matrix of $\tilde{M}$ and $k=0,\dots , K_\text{prop}-1$. Teleport probability $\alpha$ controls the balance between the Transformer features $\tilde{\mathbf{F}}$ and the propagated information. The final node representations $\mathbf{H}^{(K_\text{prop})}$ serve as graph-refined features for the classifier. 

\noindent\textbf{Classification head: }
We aggregate node features by global spatio-temporal average pooling, followed by a linear classifier: 
\begin{equation}
\tilde{\mathbf{H}} = \frac{1}{N} \sum_{i=1}^{N} \mathbf{H}_i^{(K_{\text{prop}})}, \quad \tilde{\mathbf{y}}=\text{softmax}(W_\text{cls}\tilde{\mathbf{H}}+b_\text{cls})
\end{equation}
where $ \mathbf{H}_i^{(K_\text{prop})} \in \mathbf{H}^{(K_\text{prop})}$. $W_\text{cls}$ and $b_\text{cls}$ are the learnable weight matrix and bias.



\section{Experiments}
\noindent\textbf{Datasets:} We evaluate the proposed framework on three standard fine-grained action benchmarks.
\textbf{Gym99}~\cite{finegym} contains $\mathbf{26K}$ training and $\mathbf{8.5K}$ evaluation videos covering 99 fine-grained gymnastics actions. The dataset is curated to reduce background bias, encouraging models to rely on temporal motion patterns rather than static context. \textbf{Gym288}~\cite{finegym} extends Gym99 to 288 categories with a pronounced long-tailed distribution, comprising $\mathbf{29K}$ training and $\mathbf{9.6K}$ evaluation videos. This setting is more challenging due to subtle inter-class differences and significant class imbalance. \textbf{HAA500}~\cite{haa500} includes $\mathbf{10K}$ RGB clips annotated with 500 human-centric actions. It is split into  $\mathbf{8K}$ training,  $\mathbf{0.5K}$ validation, and   $\mathbf{1.5K}$ testing clips. Unless stated otherwise, we follow the official splits and report Top-1 accuracy under the single-clip evaluation protocol described in \textit{Implementation details}.

\noindent\textbf{Implementation Details:}  
We adopt an R(2+1)D-34 backbone~\cite{r2plus1d} pre-trained on IG-65M~\cite{ig65m}. In addition, we use a 2-layer Transformer encoder (hidden size 1024, $H=8$ heads) to model long-range dependencies. All experiments are implemented in PyTorch and run on an NVIDIA RTX 6000 Ada Generation GPU. We follow the UniView strategy to uniformly sample frames from each video for both training and evaluation. Sampled frames for HAA500 and FineGym (Gym99 and Gym288) are 32 and 64, respectively. During training, each frame is resized to a short side of 126 pixels, then randomly cropped to $112\times 112$ with a random horizontal flip. At test time, we use a deterministic centre crop of $112\times 112$. Models are trained with Adam (initial learning rate $1 \times 10^{-5}$) and a cosine annealing schedule; the objective is cross-entropy. Unless otherwise stated, these settings are kept fixed across datasets.

\subsection{Comparison with State-of-the-Art (SotA)}
\noindent\textbf{GYM99 and GYM288 Datasets:}  
Table~\ref{tab:merged_comparison} shows the comparison of our \textbf{TAG-Head} with the SotA approaches using Top-1 and mean class accuracy (MCA) metrics. Our TAG-Head achieves \textbf{95.6\% Top-1 / 93.8\% MCA} on GYM99 and \textbf{92.2\% / 68.6\%} on GYM288. On GYM99, our model surpasses all RGB-only baselines and is competitive with SotA multimodal approaches. On the more demanding GYM288 (long-tailed, 288 classes), TAG-Head establishes a new SotA across both Top-1 and MCA, outperforming all published RGB-only and multimodal methods. These gains indicate that the proposed hybrid \emph{Transformer} + \emph{time-aligned graph} is highly effective at capturing subtle intra-class differences \emph{without} auxiliary modalities.

Among RGB-only models, TQN~\cite{tqn2021} was previously the strongest baseline, achieving 93.8\% Top-1 and 90.6\% MCA on GYM99, and 89.6\% Top-1 and 61.9\% MCA on GYM288. TAG-Head surpasses these results in both datasets, improving Top-1 by +1.8\% and MCA by +3.2\% in GYM99, and achieving a +2.6\% Top-1 and a substantial +6.7\% gain in MCA in GYM288. These improvements reflect a stronger performance not only in overall accuracy but also in class-balanced evaluation, particularly in the more diverse GYM288’s long tail, reflecting better calibration and coverage for rare classes. In contrast, earlier RGB-only models, such as SlowFast and TSM, trail substantially on GYM288 (Top-1: 86.8\% and 73.5\%, respectively), suggesting insufficient temporal precision and weaker fine-grained discrimination.

On GYM99, TAG-Head delivers stronger MCA than recent multimodal methods, even though \emph{only RGB} is used, with +2.2 and +2.0 over PGVT~\cite{pgvt2024} and PEVL~\cite{pevl2024}, respectively. In particular, PEVL employs pose-enhanced VL learning, relying on pose, text, and visual input to guide cross-modal representation. Although our Top-1 is slightly lower (by 1.1\% to PGVT and by 1.4\% to PEVL), the higher class-balanced accuracy indicates improved generalisation across rare/ambiguous categories. On GYM288, TAG-Head \textbf{exceeds} all multimodal entries in both metrics, including PEVL (+0.4 Top-1, +4.6 MCA), underscoring that careful spatio-temporal reasoning on RGB alone can rival or surpass the benefits of pose/text priors, particularly under class imbalance and subtle inter-class boundaries.

\begin{table}[t]
\centering
\caption{Top-1 and Mean Class Accuracy (MCA) in \% on GYM99, GYM288, and HAA500 datasets. R $\rightarrow$ RGB, T $\rightarrow$ Text, P $\rightarrow$ Pose.}
\label{tab:merged_comparison}

\begin{tabular}{|l|l|cc|cc|c|}
\hline
\textbf{Model} & \textbf{Input} 
& \multicolumn{2}{c|}{\textbf{GYM99}} 
& \multicolumn{2}{c|}{\textbf{GYM288}} 
& \textbf{HAA500 } \\
\cline{3-7}
& 
& Top-1 & MCA 
& Top-1 & MCA 
& Top-1 \\
\hline

TSN \cite{tsn2018} & R & 86.0 & 76.4 & 68.3 & 37.6 & 55.0  \\
TRNms \cite{trn2018} & R & 87.8 & 80.2 & 73.1 & 43.3 & -- \\
TSM \cite{tsm2019} & R & 88.4 & 81.2 & 73.5 & 46.5 & -- \\
SlowFast  \cite{slowfast2019} & R & 93.9 & 90.6 & 86.8 & 51.2 & 25.1 \\
I3D \cite{i3d2017} & R & 75.6 & 64.4 & 66.7 & 28.2 & 33.5 \\
TQN \cite{tqn2021} & R & 93.8 & 90.6 & 89.6 & 61.9 & -- \\
EVL \cite{frozenclip2022} & R & -- & -- & -- & -- & 75.0 \\
DC-TBAC-CSN \cite{tbac2022} & R & -- & -- & -- & -- & 83.7 \\
TPN \cite{tpn2020} & R & -- & -- & -- & -- & 50.5 \\
\hline

VTCE \cite{cct2022} & R + T & 90.1 & 91.4 & 90.1 & 62.6 & -- \\
PGVT \cite{pgvt2024} & R + P & 96.7 & 91.6 & 91.0 & 63.6 & -- \\ 
\hline

PEVL  \cite{pevl2024} & R + T + P & \textbf{97.0} & 91.8 & 91.8 & 64.0 & 84.7 \\
P2S + EVL \cite{pangea2024} & R + T + P & -- & -- & -- & -- & 80.9 \\
\hline

\textbf{ \textbf{TAG-Head (Ours)}} & \textbf{R} &  95.6 & \textbf{93.8} & \textbf{92.2} & \textbf{68.6} & \textbf{86.1} \\
\hline

\end{tabular}
\end{table}

\noindent\textbf{HAA500 Dataset:}  
Our TAG-Head reaches 86.1\% (Top-1), outperforming prior RGB-only and multimodal models (Table~\ref{tab:merged_comparison}). P2S + EVL uses multiple modalities, including RGB, 3D skeleton, and text, to improve generalisation. Despite this extensive cross-modal fusion, it achieves only 80.9\%, falling short of TAG-Head by 5.2\%. In particular, TAG-Head exceeds DC-TBAC-CSN (+2.4) and PEVL (+1.4) while avoiding their ensemble or cross-modal complexity. This suggests that our hybrid refinement not only scales to large class vocabularies but also transfers to diverse, non-sport fine-grained actions.

\subsection{Discussion: Mechanism, Limitations, and Takeaway}
\noindent \textbf{Why it works:} 
TAG-Head's gains stem from two complementary choices. First, a compact Transformer with learnable 3D positional encodings provides \emph{global}, content-adaptive context across space and time at modest cost. Second, the \emph{time-aligned} graph refines these tokens by (i) fully connecting spatial sites \emph{within} each frame to sharpen fine-grained contrasts and (ii) propagating information \emph{across} frames only along the same spatial site, which stabilises motion cues. Personalized PageRank propagation with the teleport diffusion decouples ``transform'' from ``propagate'', enabling deeper spread of evidence while the teleport term anchors features to the Transformer output, mitigating oversmoothing on dense intra-frame cliques. Together, these components recover much of the discriminative power often attributed to pose/text priors without their inference or annotation/computation overhead.

\noindent\textbf{Limitations and fairness: } 
Small Top-1 gaps to the very best multimodal models can persist on balanced settings (e.g., GYM99) where explicit pose or textual semantics help disambiguate near-duplicate classes. Our graph assumes stable spatial alignment; large camera motion may reduce temporal benefits (alleviated by stride, or light pre-alignment). We emphasise class-balanced metrics (MCA) and official splits to ensure fair comparison and avoid reliance on privileged modalities.

\noindent \textbf{Takeaway:}
The hybrid \textbf{Transformer + time-aligned graph} offers a favourable accuracy–efficiency–deployability trade-off; it matches or surpasses multimodal SotA on long-tailed FHAR (e.g., GYM288) and improves MCA substantially, all while remaining RGB-only and lightweight.

\subsection{Ablation Studies} 
In this section, we conduct extensive ablation experiments to validate our design choices and assess the individual contributions of each component within the TAG-Head framework. We specifically examine the impact of architectural hyperparameters, backbone versatility, and the synergistic effect of the transformer and graph-based refinement modules

\noindent \textbf{Impact of Transformer MLP width: }
We vary the MLP width of the Transformer (FFN dimension) to investigate the capacity–regularisation trade-off. The results in Table~\ref{tab:ffn_ablation} reveal a dataset-dependent optimum. On GYM99, FFN=1024 achieves the Top-1/MCA (95.6\% /93.8\%), outperforming 2048 by +0.9/+1.0 and 512 by +0.1/+0.5. This suggests that Gym99 benefits from a moderate feed-forward capacity to capture structured phase-like differences without overfitting. In contrast, HAA500 with broader appearance variation prefers a smaller FFN of 512 (Top-1 86.1\%), improving over 1024 and 2048 by +1.1 and +1.2, respectively, and over-compressing at 256 costs 0.5\%. This suggests a medium FFN width is sufficient for FineGym, and a smaller FFN regularises better for HAA500’s diversity.

\begin{table}[t]
\centering

\begin{minipage}{0.48\textwidth}
\centering
\caption{Impact of Transformer's MLP width (FFN) on performance using GYM99 and HAA500 datasets.}
\label{tab:ffn_ablation}
\begin{tabular}{|c|cc|c|}
\hline
\textbf{FFN Dim} & \multicolumn{2}{c|}{\textbf{GYM99 }} & \textbf{HAA500} \\
\cline{2-3}
 & Top-1 & MCA & Top-1 \\
\hline
2048 & 94.7 & 92.8 & 84.9 \\
1024 & \textbf{95.6} & \textbf{93.8} & 85.0 \\
512  & 95.5 & 93.3 & \textbf{86.1} \\
256  & 94.4 & 92.8 & 85.6 \\
\hline
\end{tabular}
\end{minipage}
\hfill
\begin{minipage}{0.48\textwidth}
\centering
\caption{Impact of Backbone Choice on HAA500 and GYM99 performance. All models use Kinetics-400 pretrained weights.}
\label{tab:backbone_ablation}
\begin{tabular}{|l|c|c|c|}
\hline
\multirow{2}{*}{\textbf{Backbone}} & \textbf{HAA500} & \multicolumn{2}{c|}{\textbf{GYM99}} \\
\cline{2-4}
 & Top-1 & Top-1 & MCA \\
\hline
I3D & 70.2 & 89.9 & 85.0 \\
I3D + TAG-Head & \textbf{72.0} & \textbf{90.2} & \textbf{85.7} \\
Slow (SlowFast) & 70.8 & 90.0 & 84.9 \\
Slow + TAG-Head & \textbf{72.7} & \textbf{90.3} & \textbf{85.8} \\
CSN  & 76.1 & 89.4 & 85.1 \\
CSN +TAG-Head & \textbf{77.5} & \textbf{90.0} & 85.1 \\
\hline
\end{tabular}
\end{minipage}

\end{table}

\noindent \textbf{Effect of Backbone Architecture:} To demonstrate the ``plug-and-play'' versatility of TAG-Head, we evaluated its performance when integrated with several standard video backbones: I3D, the Slow pathway of  SlowFast-R50, and  CSN. For comparison, all backbones were initialised with Kinetics-400 pretrained weights. As summarised in Table~\ref{tab:backbone_ablation}, the addition of TAG-Head consistently enhances performance across both benchmarks, regardless of the underlying feature extractor. On HAA500, TAG-Head delivers consistent Top-1 gains: +1.8\% (I3D), +1.9\% (Slow), and +1.4\% (CSN). Similar trends appear on GYM99, where it improves the MCA of the Slow pathway and I3D by +0.9\% and +0.7\%, respectively. For CSN, the module improves Top-1 accuracy from 89.4\% to 90.0\%. These results confirm that TAG-Head is backbone-agnostic, successfully capturing the nuanced spatio-temporal dependencies, local spatial interactions and stable temporal propagation that standard 3D CNNs often fail to resolve.

\begin{table}[t]
\centering
\caption{Ablation study on HAA500 and GYM288. Accuracy is reported as Top-1 (\%). B: 3D Backbone, TE: Transformer Encoder, IF-FC: Intra-frame Fully Connected edges, TAT: Time-aligned Temporal edges.}
\label{tab:ablation-comparison}
\begin{tabular}{@{}lcc@{}}
\toprule
\textbf{Configuration} & \textbf{HAA500} & \textbf{GYM288} \\
\midrule
B                                 & 83.2          & 89.7          \\
B + TE                            & 84.9          & 91.4          \\
B + TE + IF-FC                    & 84.4          & 92.0          \\
B + TE + TAT                     & 85.1          & 90.4          \\
\textbf{Full (B + TE + IF-FC + TAT)} & \textbf{86.1} & \textbf{92.2} \\
\bottomrule
\end{tabular}
\end{table}

\noindent \textbf{Effectiveness of Transformer and Graph-based Feature Refinement:} 
To evaluate the TAG-Head components, we conduct an ablation study by incrementally integrating the Transformer Encoder (TE), intra-frame fully connected (IF-FC), and time-aligned (TAT) graph edges into the 3D backbone ($\mathcal{B}$). As shown in Table~\ref{tab:ablation-comparison}, adding TE improves Top-1 accuracy by 1.7\% on both HAA500 (84.9\%) and GYM288 (91.4\%), validating the importance of modelling global dependencies across the token sequence. Integrating individual graph components yields dataset-specific behaviours. IF-FC improves the performance on GYM288 but slightly degrades the results on HAA500, while TAT enhances the performance on HAA500 but reduces the performance on GYM288. These fluctuations indicate that isolated spatial refinements or temporal constraints may introduce noise depending on the underlying action characteristics. Ultimately, the full configuration ($\mathcal{B} + \text{TE} + \text{IF-FC} + \text{TAT}$) achieves the best performance, reaching 86.1\% on HAA500 and 92.2\% on GYM288. This result validates the synergy between the Transformer’s global contextual modelling and graph-based diffusion, demonstrating that the joint integration of spatial contrast and temporal continuity is critical for fine-grained action recognition.

\begin{figure}[H] 
    \centering
    \begin{subfigure}[t]{0.31\linewidth}
        \centering
        \includegraphics[width=\linewidth]{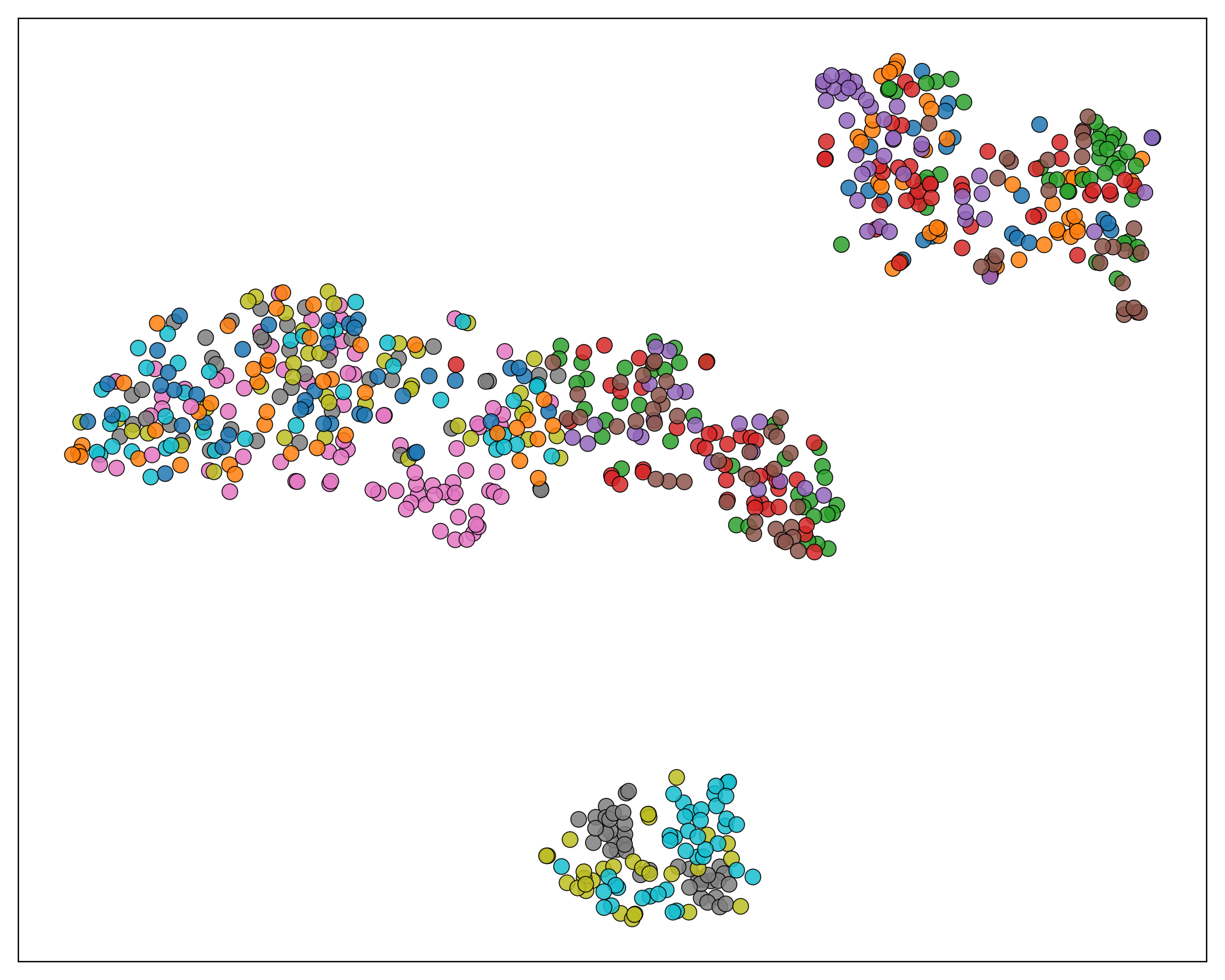}
        \caption{Gym99: TL}
    \end{subfigure}
    \hfill
    \begin{subfigure}[t]{0.31\linewidth}
        \centering
        \includegraphics[width=\linewidth]{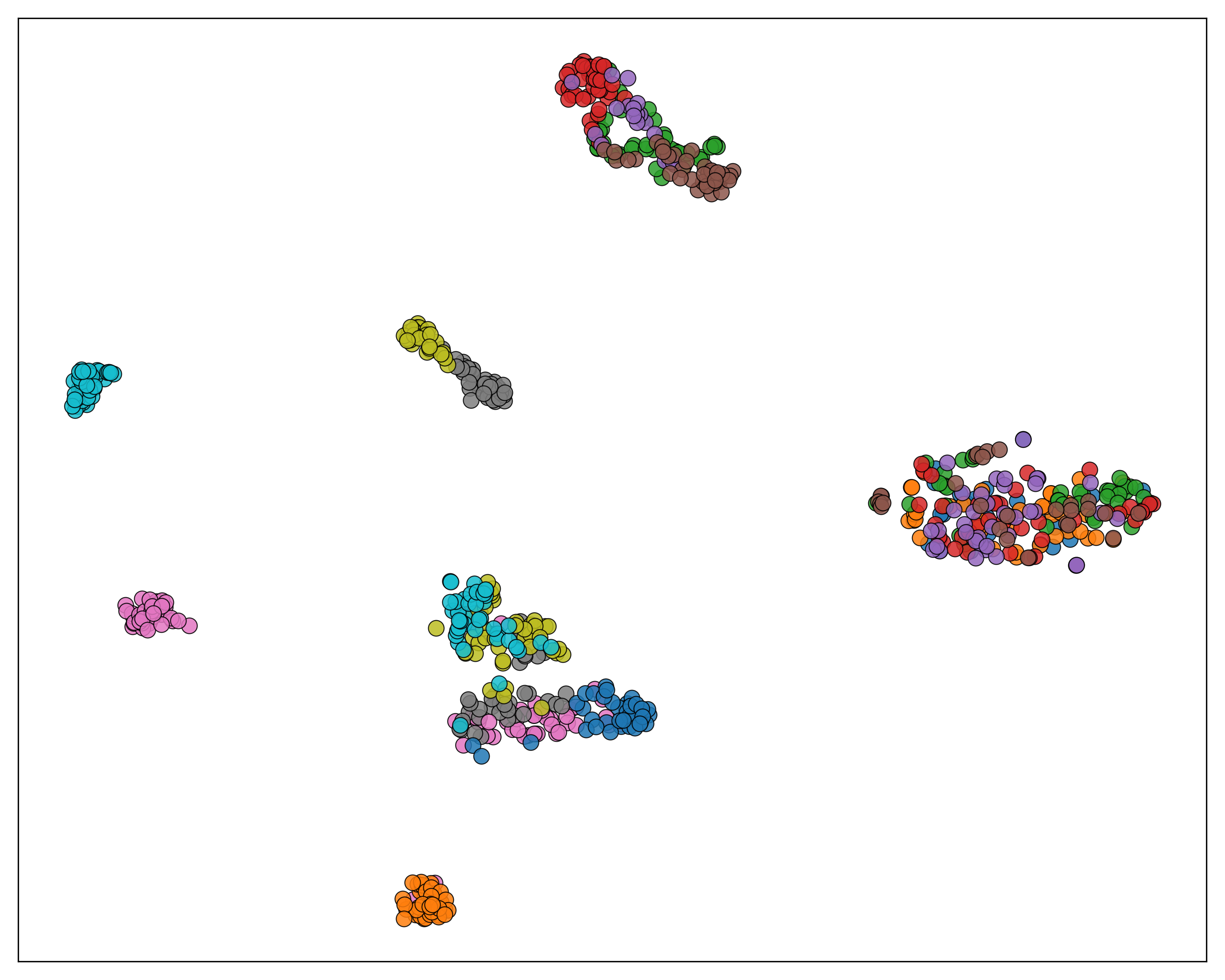}
        \caption{Gym99: FT}
    \end{subfigure}
    \hfill
    \begin{subfigure}[t]{0.31\linewidth}
        \centering
        \includegraphics[width=\linewidth]{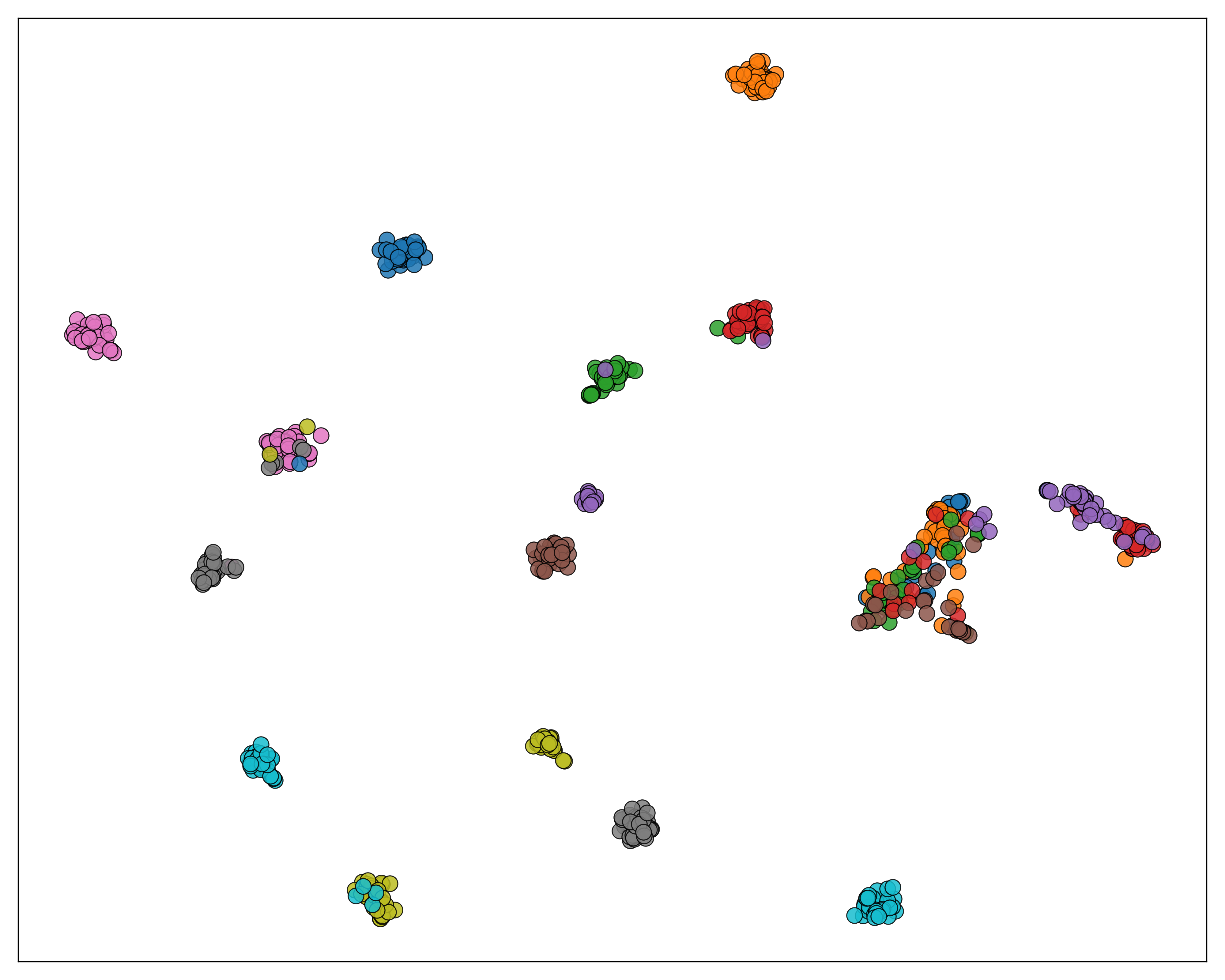}
        \caption{Gym99: TAG-Head}
    \end{subfigure}

    \vspace{0.3em} 

    \begin{subfigure}[t]{0.31\linewidth}
        \centering
        \includegraphics[width=\linewidth]{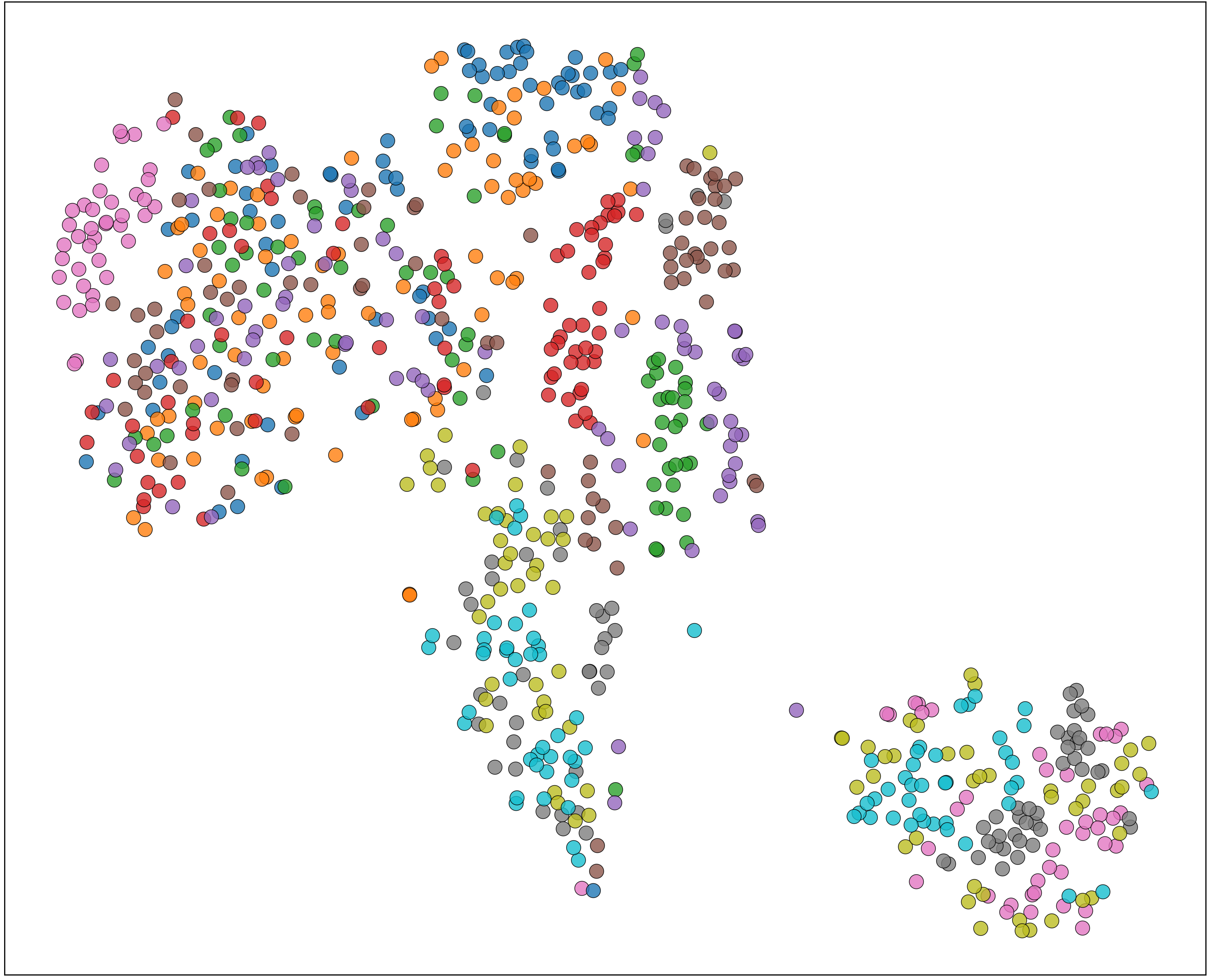}
        \caption{Gym288: TL}
    \end{subfigure}
    \hfill
    \begin{subfigure}[t]{0.31\linewidth}
        \centering
        \includegraphics[width=\linewidth]{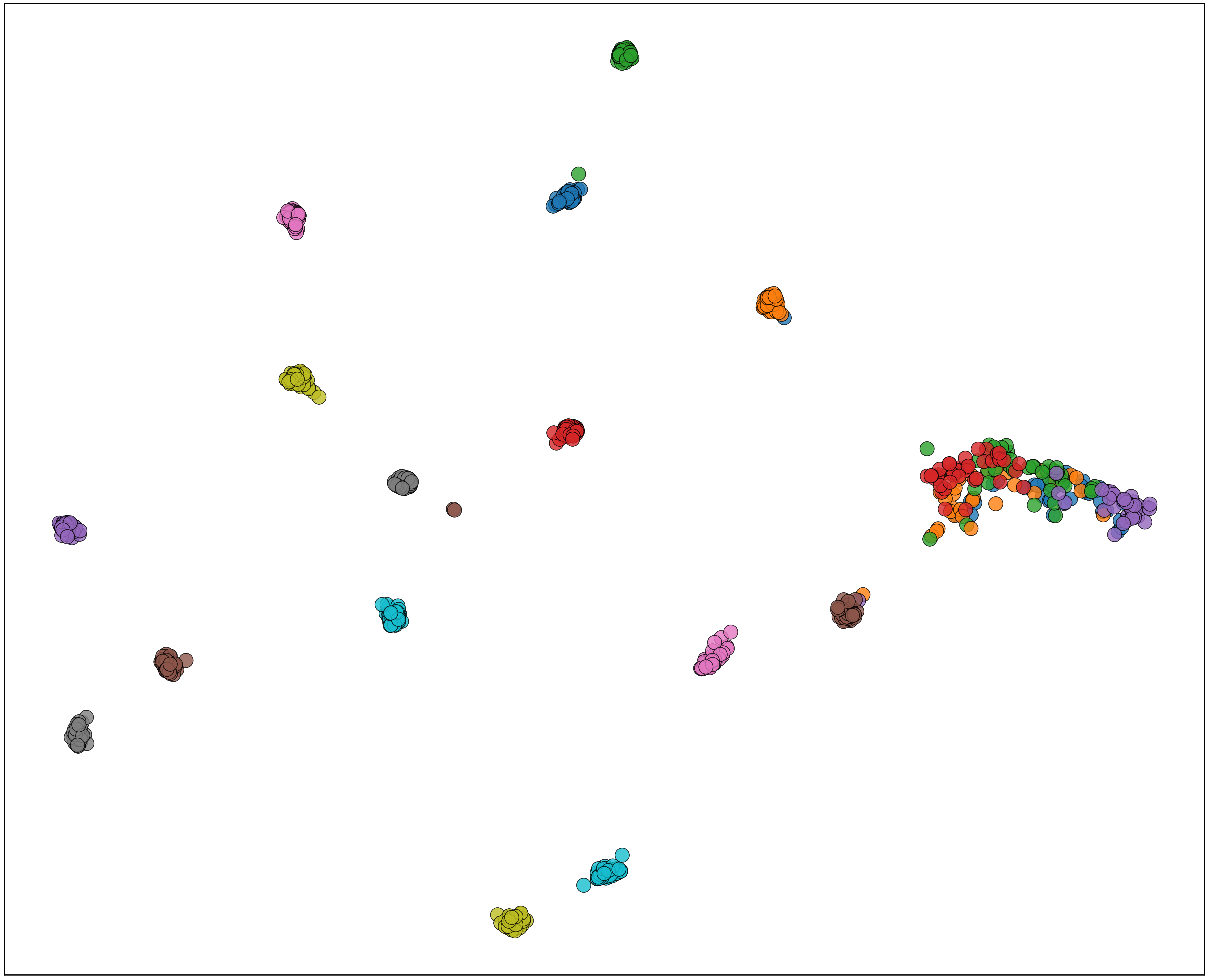}
        \caption{Gym288: FT}
    \end{subfigure}
    \hfill
    \begin{subfigure}[t]{0.31\linewidth}
        \centering
        \includegraphics[width=\linewidth]{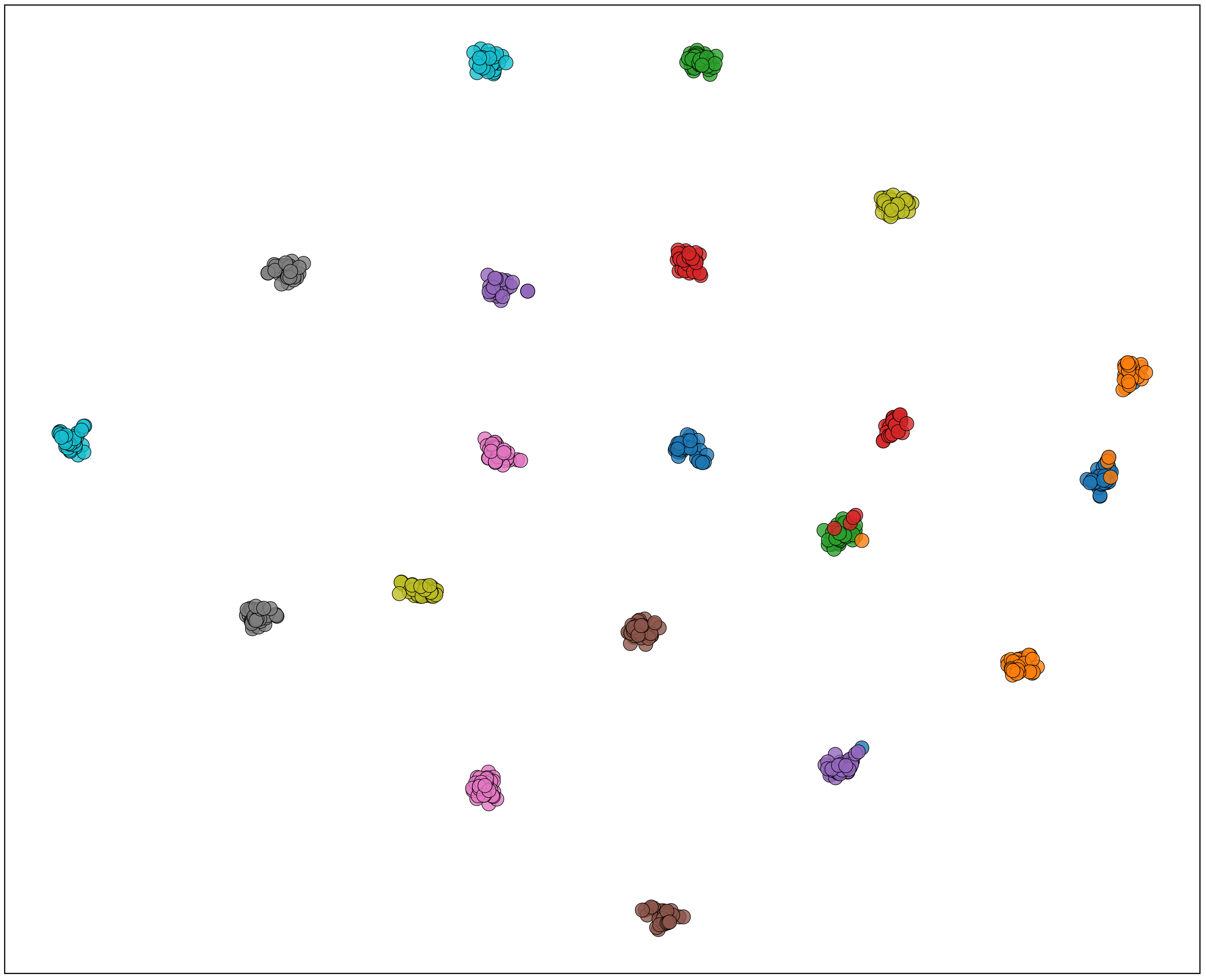}
        \caption{Gym288: TAG-Head}
    \end{subfigure}

    \vspace{0.3em}

    \begin{subfigure}[t]{0.31\linewidth}
        \centering
        \includegraphics[width=\linewidth]{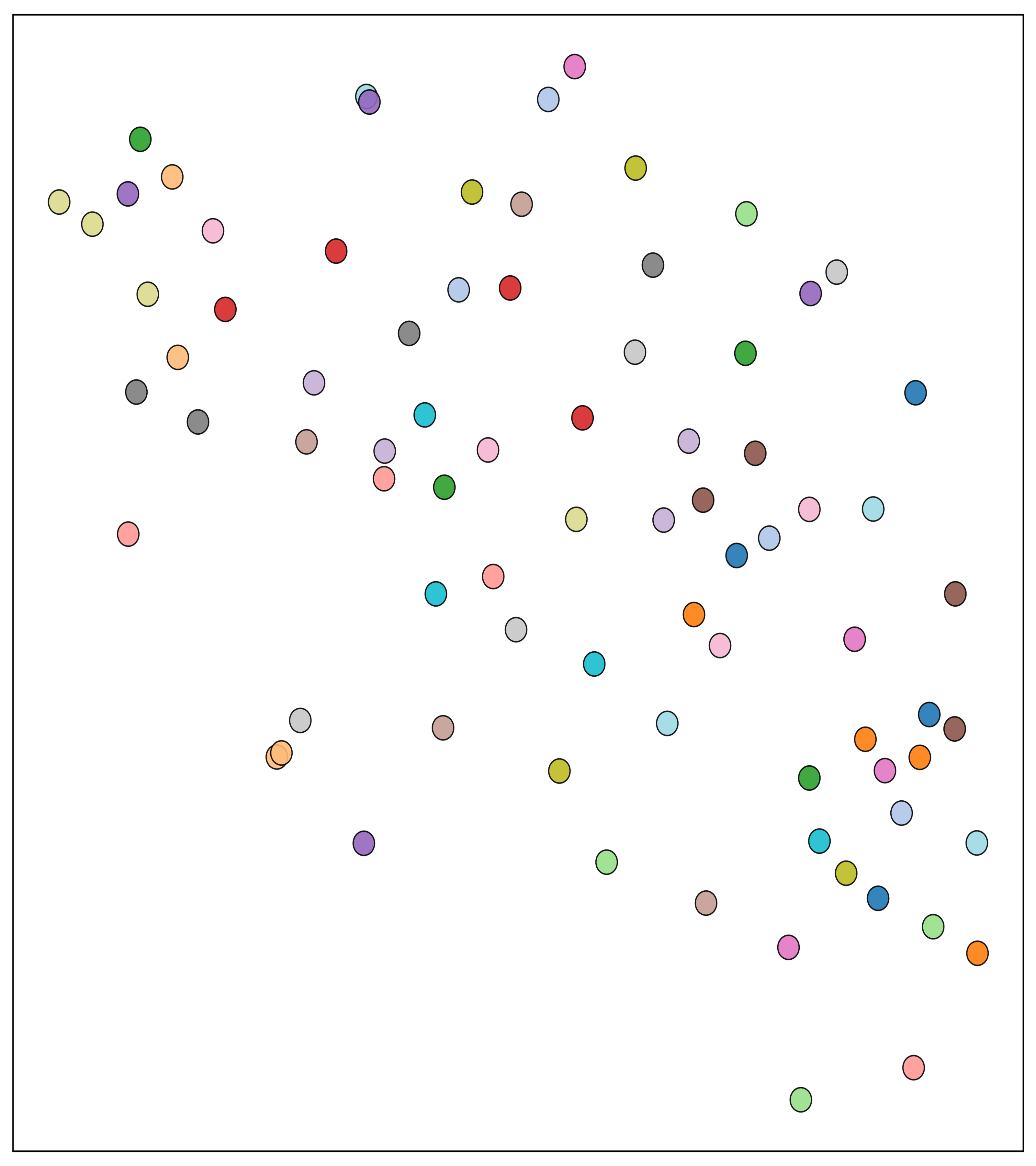}
        \caption{HA500: TL}
    \end{subfigure}
    \hfill
    \begin{subfigure}[t]{0.31\linewidth}
        \centering
        \includegraphics[width=\linewidth]{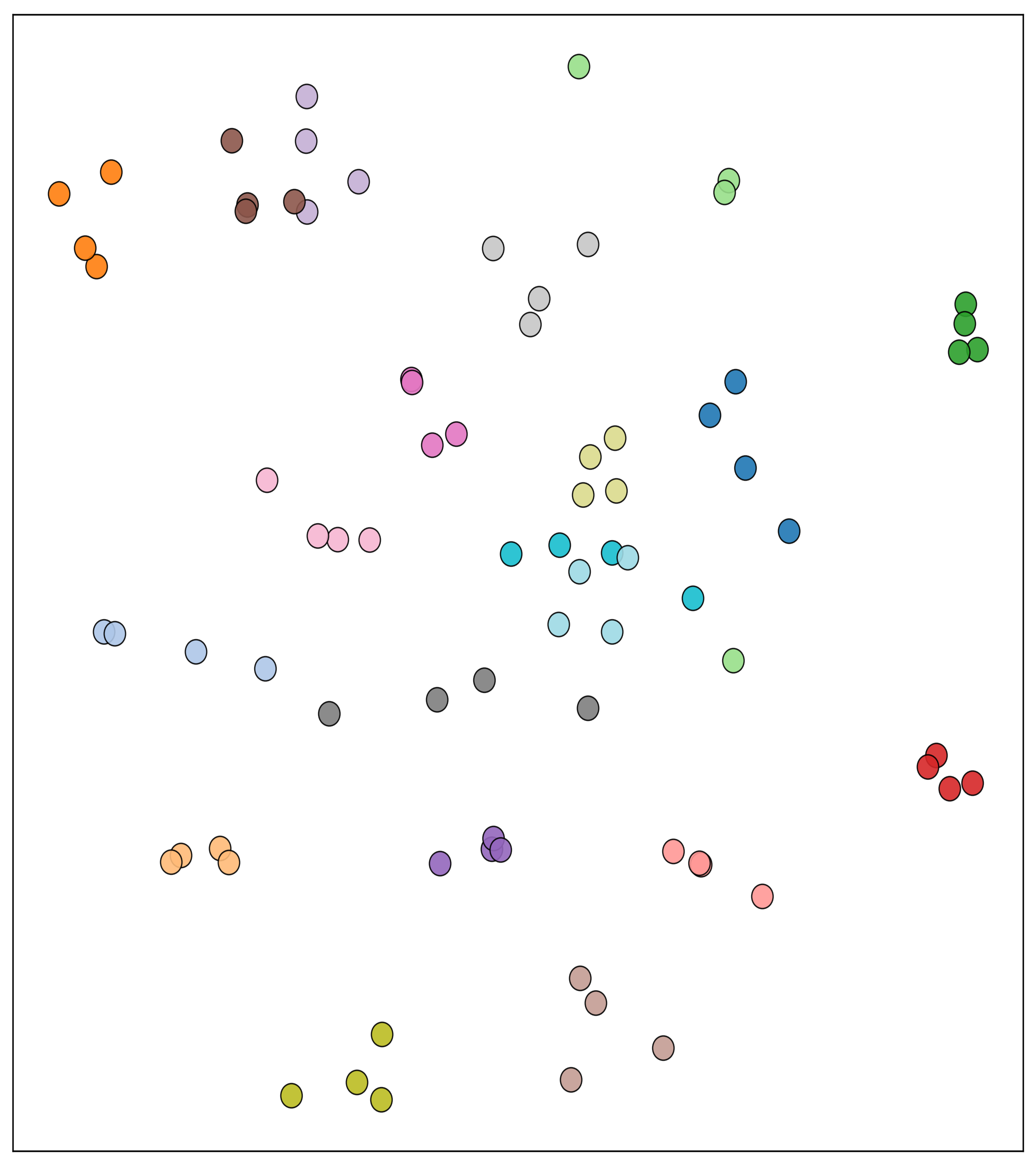}
        \caption{HA500: FT}
    \end{subfigure}
    \hfill
    \begin{subfigure}[t]{0.31\linewidth}
        \centering
        \includegraphics[width=\linewidth]{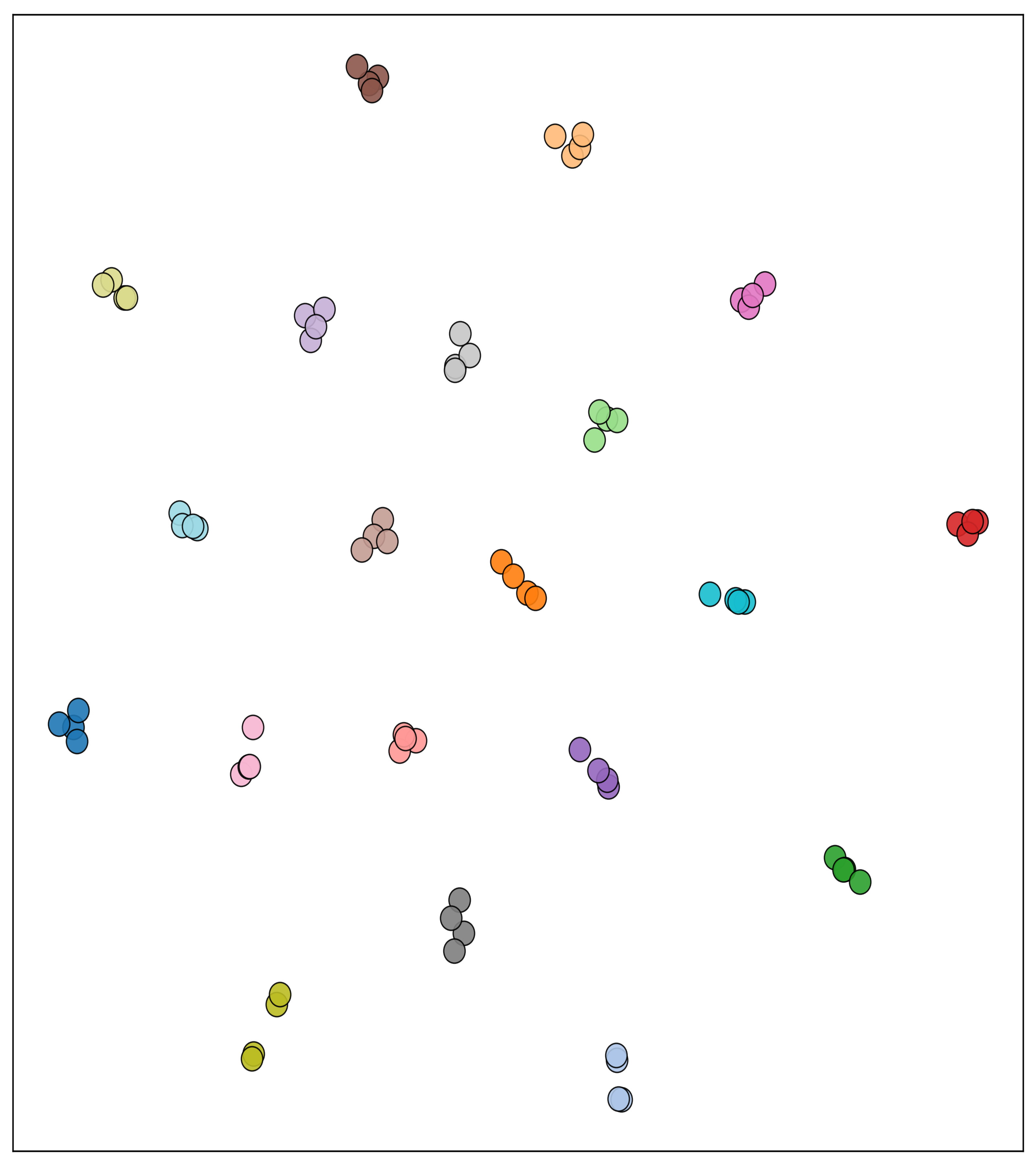}
        \caption{HA500: TAG-Head}
    \end{subfigure}

    \caption{t-SNE visualisations across Gym99, Gym288, and HA500. Columns represent: Transfer Learning (TL), Fine-tuning (FT), and the proposed TAG-Head. (a, d, g) TL features lack separation. (b, e, h) FT improves clustering. (c, f, i) TAG-Head yields the most distinct separation across all datasets.}
    \label{fig:tsne-comprehensive}
\end{figure}

\subsection{Qualitative Analysis: Feature Embedding Visualization}
To visually assess the discriminative power of the proposed TAG-Head, we analyse the feature embeddings using t-SNE across the Gym99, Gym288, and HAA500 datasets. Fig.~\ref{fig:tsne-comprehensive} compares three stages of representation: Transfer Learning (TL) with a frozen backbone, standard Fine-tuning (FT), and our proposed TAG-Head. In the TL stage (Fig.~\ref{fig:tsne-comprehensive}a, d, g), features lack clear separation, reflecting the gap between general video pre-training and FHAR requirements. While FT (Fig.~\ref{fig:tsne-comprehensive}b, e, h) improves clustering, many classes remain entangled. The TAG-Head (Fig.~\ref{fig:tsne-comprehensive}c, f, i) yields the most distinct separation and cohesive clusters across all datasets. This confirms that the complementary interaction between the Transformer's global context and the Graph-based feature Refinement successfully resolves subtle inter-class boundaries, producing a more discriminative manifold for fine-grained recognition.

\subsection{Computational Complexity and Efficiency}
To assess the practical deployability of TAG-Head, we compare its computational requirements against recent state-of-the-art multimodal methods. As summarised in Table~\ref{tab:complexity_combined} (left), TAG-Head achieves superior performance with a significantly smaller footprint. It requires only 69.90M parameters, which is approximately 3.8$\times$ fewer than PGVT~\cite{pgvt2024} and 15.7$\times$ fewer than PEVL~\cite{pevl2024}. 

The architectural breakdown in Table~\ref{tab:complexity_combined} (right) highlights the efficiency of our modular design. While the R(2+1)D-34 backbone constitutes the bulk of the parameters (63.49M), the task-specific TAG-Head components add only 6.41M parameters in total. Notably, the graph-based feature propagation using APPNP is entirely parameter-free, enabling deep relational reasoning without increasing the model's learnable capacity.

\begin{table}[t]
    \centering
    \caption{Efficiency Analysis. Left: Comparison with state-of-the-art multimodal models. Right: Internal parameter breakdown of the TAG-Head framework.}
    \label{tab:complexity_combined}
    \begin{minipage}{0.52\textwidth}
        \centering
        \begin{tabular}{l c c}
            \toprule
            \textbf{Model} & \textbf{Params (M)} & \textbf{GFLOPs} \\
            \midrule
            PGVT~\cite{pgvt2024} & 265.0 & 222.0 \\
            PEVL~\cite{pevl2024} & 1095.0 & 510.0 \\
            \textbf{TAG-Head (Ours)} & \textbf{69.90} & \textbf{153.4} \\
            \bottomrule
        \end{tabular}
    \end{minipage}
    \hfill
    \begin{minipage}{0.45\textwidth}
        \centering
        \begin{tabular}{l r}
            \toprule
            \textbf{Component} & \textbf{Params (M)} \\
            \midrule
            Backbone & 63.492 \\
            TAG-Head & 6.405 \\
            \bottomrule
        \end{tabular}
    \end{minipage}
\end{table}

\section{Conclusion}
In this paper, we introduced \textbf{TAG-Head}, a lightweight and modular spatio-temporal graph head designed to advance RGB-only FHAR. By strategically coupling a Transformer encoder for global context modelling with a specialised, time-aligned GNN for local refinement, our approach successfully captures the nuanced spatio-temporal dynamics essential for distinguishing visually similar actions. Unlike many recent state-of-the-art systems that rely on privileged modalities such as pose or text, TAG-Head achieves superior discriminability using only RGB data, significantly reducing annotation burdens and computational overhead.

Extensive evaluations on the FineGym (Gym99 and Gym288) and HAA500 benchmarks demonstrate that TAG-Head consistently outperforms existing RGB-only baselines and surpasses many multimodal frameworks. Notably, on the long-tailed Gym288 dataset, our model achieves a substantial gain in Mean Class Accuracy (MCA), highlighting its ability to generalise across rare and ambiguous categories. Qualitative t-SNE visualisations further validate that the hybrid Transformer-Graph architecture produces more distinct and separable feature clusters than standard fine-tuning.

The compact, plug-and-play nature of TAG-Head ensures it can be easily integrated into existing 3D backbones with minimal parameter overhead. By delivering performance typically associated with heavier multimodal models, TAG-Head provides a scalable and efficient solution for real-world FHAR applications, such as sports analytics and surveillance, where sensor simplicity and low latency are paramount.

\noindent\textbf{Acknowledgements: }
This research is supported by the UK Research and Innovation (UKRI) - Engineering and Physical Science Research Council (EPSRC) under under Grant EP/X028631/1 (ATRACT project).

%
%
%
\bibliographystyle{splncs04}
\bibliography{mybibliography}

\end{document}